%% file: main.tex
\documentclass[11pt,letterpaper]{article}
\usepackage{PRIMEarxiv}
\input{preamble}
\newcommand{\method}{ViLFlow\xspace}
\title{\method: Parallel Vision–Language Generation with Continuous Latent Flows}
\author{Kevin Zhai\thanks{Equal contribution.} \quad Bian Sun\footnotemark[1] \quad Zhenyi Wang\footnotemark[2] \quad Song Wang\thanks{Correspondence to \href{mailto:song.wang@ucf.edu}{song.wang@ucf.edu}}\\University of Central Florida\\Institute of Artificial Intelligence}
\date{}
\hypersetup{pdftitle={ViLFlow: Parallel Vision–Language Generation with Continuous Latent Flows},pdfauthor={Kevin Zhai, Bian Sun, Zhenyi Wang, Song Wang},pdfsubject={Continuous vision-language generation},pdfkeywords={vision-language models,continuous diffusion,visual question answering,image captioning}}
\begin{document}
\maketitle
\input{sections/abstract}

\input{sections/introduction}
\input{sections/related_work}
\input{sections/formulation}

\input{sections/method2}
\input{sections/experiments_current}
\input{sections/conclusion}

\end{document}

%% file: preamble.tex
\usepackage[utf8]{inputenc}
\usepackage[T1]{fontenc}
\usepackage{amsmath,amssymb}
\usepackage[dvipsnames]{xcolor}
\usepackage{graphicx}
\usepackage{booktabs,array,tabularx}
\usepackage[numbers,sort&compress]{natbib}
\usepackage[hyphens]{url}
\usepackage{caption}
\usepackage{subcaption}
\usepackage{placeins}
\usepackage{float}
\usepackage{tikz}
\usetikzlibrary{arrows.meta,positioning}
\usepackage{fontawesome5}
\usepackage{xspace}
\usepackage{enumitem}
\usepackage{etoolbox}
\makeatletter
\patchcmd{\normalsize}{\@xpt\@xipt}{\@xipt{13}}{}{\PackageError{arxiv-layout}{Body font patch failed}{Check the pinned PRIME template.}}
\patchcmd{\small}{\@ixpt\@xpt}{\@xpt\@xiipt}{}{\PackageError{arxiv-layout}{Small font patch failed}{Check the pinned PRIME template.}}
\patchcmd{\footnotesize}{\@ixpt\@xpt}{\@ixpt{11}}{}{\PackageError{arxiv-layout}{Footnote font patch failed}{Check the pinned PRIME template.}}
\normalsize
\patchcmd{\@maketitle}{\vskip 0.4in \@minus 0.1in \center{ } \vskip 0.2in}{\vskip 0.18in}{}{\PackageError{arxiv-layout}{Title spacing patch failed}{Check the pinned PRIME template.}}
\makeatother
\definecolor{linkblue}{rgb}{0.21,0.49,0.74}
\usepackage[breaklinks,colorlinks,allcolors=linkblue]{hyperref}
\usepackage[capitalize]{cleveref}

%% file: sections/abstract.tex
\begin{abstract}

Vision-language generation has been predominantly approached through autoregressive models and discrete diffusion language models. However, autoregressive models suffer from inefficiencies in sequential decoding, while discrete diffusion models typically predict tokens independently at each denoising step, limiting their ability to explicitly capture dependencies among generated tokens. This raises a fundamental question: How can we achieve efficient parallel vision-language generation while preserving the joint dependencies among output tokens and effectively incorporating visual information? To address this question, we propose \method (Vision--Language Flow), a continuous-space generative framework that extends pretrained continuous language generation to vision-language tasks, including visual question answering and image captioning. By modeling the joint distribution of token embeddings through continuous flow-based generation, \method enables parallel decoding while capturing dependencies across output positions. To effectively incorporate visual semantics, we introduce a teacher-guided distillation strategy that transfers image-conditioned knowledge into the denoising process, enabling visual grounding with a single image-conditioned prediction at each generation step. Furthermore, to bridge the discrepancy between continuous denoising outputs and discrete linguistic representations, we develop a prediction-aware token decoder trained using both denoiser predictions and paired reference text. This decoder can be further adapted independently of the denoiser, improving the quality of generated text without additional denoiser training. \method outperforms autoregressive and masked-diffusion baselines by up to 8.46 percentage points on VQAv2 and 4.50 points on GQA without GQA training. For image captioning, \method improves CIDEr by 16.25 over masked diffusion and 17.81 over AR baselines.

\end{abstract}

%% file: sections/introduction.tex
\section{Introduction}
\label{sec:intro}

Vision--language models have conventially followed two paradigms. Autoregressive models~\citep{liu2023llava} sequentially generate tokens left-to-right, which limits parallel decoding. More recent discrete diffusion models~\cite{you2026lladav,li2025lavida,yang2025mmada}, despite generating tokens in parallel, must still ensure these tokens express mutually consistent and coherent responses. However, masked diffusion typically predicts tokens independently within each denoising step, conditioned on the current sequence~\cite{sahoo2024mdlm}. This raises the question: how can vision--language models jointly model output dependencies while retaining parallel decoding and incorporating visual information?

Continuous language generation offers a way to combine joint sequence modeling with parallel decoding~\cite{hu2026elf}. Rather than unmasking tokens, it progressively refines a continuous representation of the entire response before converting that representation into text. This allows different parts of the response to develop together while retaining parallel generation. For vision--language generation, visual information must guide this joint refinement and carry through to the final text. Hence, visual adaptation requires addressing two key aspects: how the model generates continuous predictions and how it learns to decode those predictions into accurate answers and captions.

\input{figures/overview}

We thus introduce \method (Vision--Language Flow), a continuous-flow vision--language model that adapts pretrained language generation through learned image guidance and prediction-aware token decoding (Figure~\ref{fig:overview}). Alongside reference text, \method uses model predictions to train both image-conditioned denoising and token decoding. To learn image guidance, a visually adapted teacher predicts the same noisy text with and without image features. We amplify the difference between these predictions to construct a guided training target~\cite{meng2023distillation}. By learning to reproduce this target, the denoiser incorporates image guidance into a single image-conditioned prediction at each generation step. Computing the same guidance during inference would require two predictions.

Adapting the token decoder improves both answering and captioning without changing the trained denoiser's weights. The decoder initially learns to reconstruct text from corrupted reference embeddings, whereas generation supplies denoiser predictions. We therefore train the decoder on these predictions to improve their conversion into words. This adapted decoder also supplies text feedback during late denoising, using its clean predictions to refine the continuous response before final decoding.

We evaluate \method against autoregressive and masked-diffusion baselines on visual question answering (VQA) and image captioning. On the shared VQAv2 evaluation set~\cite{goyal2017vqa}, \method achieves 63.39\% accuracy, exceeding our GPT-2-based autoregressive baseline by 3.34 percentage points. It also generalizes to GQA~\cite{hudson2019gqa} without further training, achieving 39.47\% accuracy and a 2.83-point improvement over the autoregressive baseline. For captioning, \method achieves 111.16 CIDEr on the COCO Karpathy test split~\cite{lin2014coco,karpathy2015deep}, outperforming our 32-step masked-diffusion baseline by 16.25 points. Our contributions are as follows:

\begin{itemize}[leftmargin=1.1em,labelsep=0.4em]
    \setlength{\itemsep}{0.5em}
    \setlength{\parskip}{0pt}
    \setlength{\parsep}{0pt}
    \interlinepenalty=10000

\item We introduce \method, a continuous-flow vision--language model that adapts pretrained language generation to visual tasks through joint embedding generation and parallel token decoding.

\item We distill image guidance from a visually adapted teacher into the denoiser, enabling image-guided generation with one image-conditioned prediction per step.

\item We develop prediction-aware token decoding that improves answering and captioning without further denoiser training. The adapted decoder also provides text feedback during caption generation.

\item We show that \method outperforms our autoregressive baseline on VQA and generalizes to GQA without further training. Captioning gains over masked diffusion are complemented by controlled evidence that our training choices improve performance beyond additional training alone.

\end{itemize}

%% file: figures/overview.tex
\begin{figure}[H]
\centering
\includegraphics[width=0.68\linewidth]{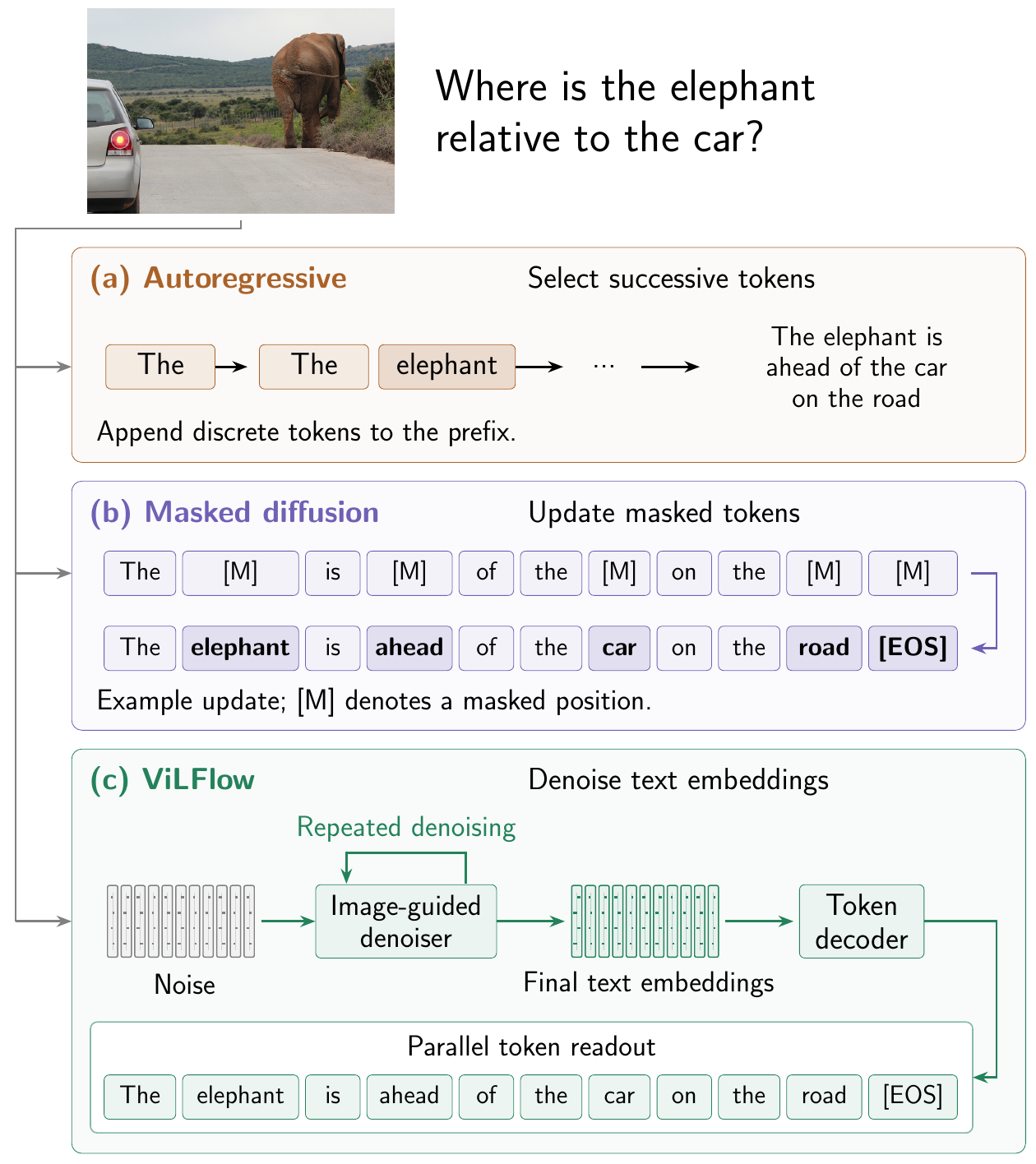}
\caption{\textbf{Three approaches to vision-language generation, illustrated with the same image and question.} (a) Autoregressive models decode left-to-right. (b) Masked diffusion models replace masks [M] with token predictions, highlighted in the second row. (c) \method repeatedly denoises continuous text embeddings before decoding tokens in parallel. Each narrow outline represents one output position, with schematic coordinates shown inside. Gray embeddings denote initial noise and green embeddings the final state.}
\label{fig:overview}
\end{figure}

%% file: sections/related_work.tex
\section{Related Work}

\subsection{Language Models}

Autoregressive language models such as GPT-2~\cite{radford2019gpt2} and LLaMA~\cite{touvron2023llama} predict each token from the preceding sequence. Masked diffusion language models, including MDLM~\cite{sahoo2024mdlm} and LLaDA~\cite{nie2025llada}, instead recover masked tokens in parallel using bidirectional context. Both approaches represent the evolving output as discrete tokens or masks.

Continuous language models differ in their text representations and token decoding. Earlier approaches explore continuous text denoising for controlled and conditional generation~\cite{li2022diffusionlm,gong2023diffuseq,lin2023genie,strudel2022sed,karimimahabadi2024tess}. CoLa~\cite{guo2026cola} learns a text variational autoencoder and a block-causal continuous prior, with preliminary experiments on image-conditioned generation. ELF~\cite{hu2026elf} instead denoises embeddings from a frozen contextual text encoder and decodes tokens after denoising. ELF-S2T~\cite{li2026elfs2t} applies this formulation to speech. Sigma~\cite{yang2026sigma} uses vocabulary probabilities to construct embedding predictions during denoising, while HC-DLM~\cite{ren2026hcdlm} uses provisional tokens as inputs to subsequent continuous updates. \method adopts ELF's frozen text representation and pretrained full-sequence denoiser as the foundation for visual adaptation.

\subsection{Vision-Language Models}

LLaVA~\cite{liu2023llava} adapts autoregressive language modeling to visual tasks through a learned projector that maps image features into the language model's embedding space. Other pretrained VLMs connect vision and language models through learned interfaces~\cite{li2023blip2,dai2023instructblip,alayrac2022flamingo}. LLaDA-V~\cite{you2026lladav}, LaViDa~\cite{li2025lavida}, and MMaDA~\cite{yang2025mmada} instead condition masked-token generation on images. Discrete diffusion has also been applied to captioning-specific and unified multimodal models~\cite{zhu2022ddcap,li2025dualdiffusion}. For text generation, these models represent the evolving output with discrete tokens or masks. \method denoises continuous text embeddings and decodes the output tokens in parallel at the final readout.

Continuous diffusion has also been studied directly for image captioning. Analog Bits and SCD-Net apply continuous diffusion to bit representations of caption tokens~\cite{chen2023analogbits,luo2023scdnet}. DiffCap~\cite{he2023diffcap} denoises learned caption embeddings, and LaDiC~\cite{wang2024ladic} uses a split BERT encoder and decoder around an image-conditioned diffuser. UniDiffuser~\cite{bao2023unidiffuser} studies unified image--text diffusion, while Multimodal Flow~\cite{tao2026multimodalflow} and JUICE~\cite{li2026juice} use continuous flows in pretrained representation spaces for unified multimodal modeling. \method adapts a pretrained continuous language generator, whereas Multimodal Flow and JUICE train their flow backbones from scratch. \method combines learned image guidance with a separate token decoder trained on predictions from the frozen denoiser. The adapted decoder also supplies text feedback during late caption-denoising updates. Matched controls separate the effects of guidance supervision, decoder inputs and inference feedback.

\input{figures/architecture}

%% file: figures/architecture.tex
\begin{figure*}[t]
\centering
\includegraphics[width=\textwidth]{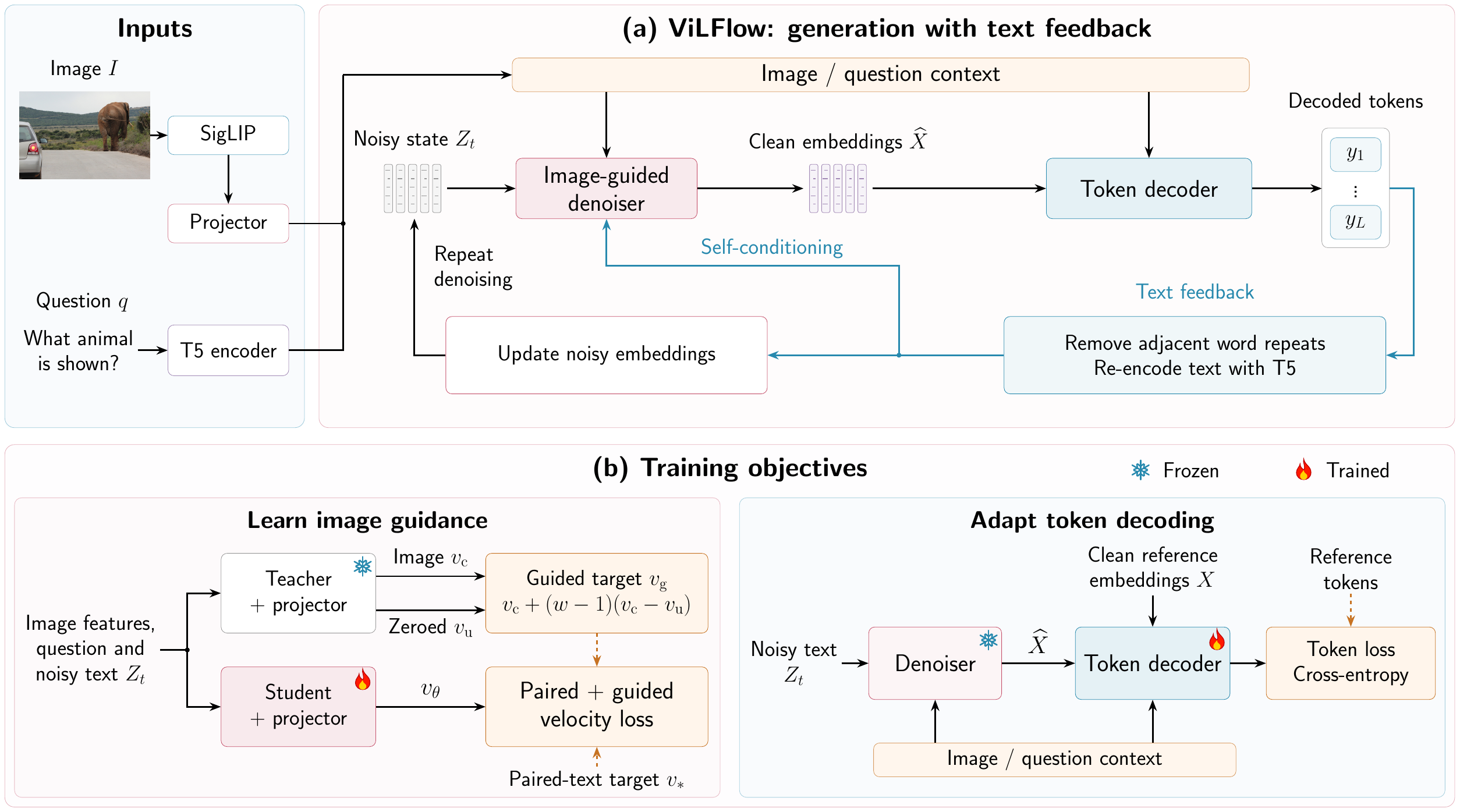}
\caption{\textbf{Overview of \method for parallel vision--language generation through continuous latent flow.} (a) Starting from Gaussian noise, the image-guided denoiser jointly refines all output positions. At selected late captioning steps, decoded text undergoes adjacent-word repetition removal and is re-encoded by frozen T5 for the current update and subsequent self-conditioning; other steps use the denoiser prediction directly. Final tokens are decoded in parallel. (b) Left: image-conditioned and image-zeroed teacher predictions provide a guided target for student training alongside paired-text supervision. Right: a separate decoder learns from denoiser predictions and clean reference embeddings using token cross-entropy, with the denoiser and visual projector frozen. Orange dashed arrows indicate supervision.}
\label{fig:architecture}
\end{figure*}

%% file: sections/formulation.tex
\section{Preliminaries}
\label{sec:formulation}

We adopt ELF's continuous text representation and its denoising and token-reconstruction objectives~\cite{hu2026elf}. A frozen text encoder maps a reference sequence padded to $L$ positions to normalized $d$-dimensional embeddings $X\in\mathbb{R}^{L\times d}$. Let $\epsilon\in\mathbb{R}^{L\times d}$ contain independent standard Gaussian entries and let $\sigma>0$ be the noise scale. The noisy embeddings at time $t\in[0,1]$ are
\begin{equation}
Z_t=tX+(1-t)\sigma\epsilon.
\label{eq:corruption}
\end{equation}

The denoiser, parameterized by $\theta$, uses a transformer and denoising head to predict clean embeddings $\widehat X$ from $Z_t$ at all positions in parallel. For flow matching, we convert the predicted and reference embeddings into a predicted velocity $v_\theta$ and target velocity $v_*$. Let $\delta(t)$ denote $1-t$ bounded away from zero for numerical stability. These velocities are
\begin{equation}
v_\theta=\frac{\widehat X-Z_t}{\delta(t)},\qquad v_*=\frac{X-Z_t}{\delta(t)}.
\label{eq:velocity}
\end{equation}
Let $m$ mark the supervised token positions and $\lVert\cdot\rVert_m^2$ denote mean squared error over those positions. The denoising loss averages the velocity error over training examples, sampled times and noise: $\mathcal{L}_{\mathrm{data}}=\mathbb{E}\lVert v_\theta-v_*\rVert_m^2$. For token reconstruction, the same transformer uses a vocabulary head to predict reference tokens in parallel from corrupted embeddings. We train this branch with cross-entropy.

%% file: sections/method2.tex
\section{Method}
\label{sec:method}

\method is a continuous diffusion vision-language model that generates captions or answers conditioned on images and optional questions. It combines full-sequence embedding-space denoising with parallel token decoding. Training comprises three stages: image-conditioned adaptation, image guidance distillation, and prediction-adapted decoding. For captioning, inference additionally incorporates text feedback into the denoising trajectory.

\subsection{Model Architecture}

\method performs image captioning and visual question answering through continuous diffusion in the text embedding space (Figure~\ref{fig:architecture}(a)). Alongside a frozen image encoder $F$ and a frozen text encoder $E$, the model contains a visual projector, a denoising transformer, and a token decoder. The projector maps image features into the text embedding space, and the projected features are concatenated with the encoded question, when present, to form the conditioning sequence. Starting from Gaussian noise, the denoising transformer iteratively refines all output positions jointly under this conditioning. During Stage I, denoising and token reconstruction share a transformer; Stage III initializes a separate decoder from its guided counterpart. This decoder also receives the multimodal conditioning and predicts tokens in parallel. For captioning, it is applied at selected late denoising steps: the decoded text is re-encoded by $E$ and used in the current sampler update and as self-conditioning for the next step. After the final update, the decoder produces the output text.

\subsection{Stage I: Image-Conditioned Adaptation}

Given an image--text pair $(I,y)$ and an optional question $q$, let $F$ denote the frozen image encoder and $E$ the frozen text encoder with output normalization. We introduce a learnable visual projector $P_\phi$ to map image features into the text embedding space. The multimodal conditioning sequence $C$ and target text embeddings $X$ are defined as
\begin{equation}
C = [P_\phi(F(I)); E(q)], \qquad X = E(y),
\label{eq:visual_condition}
\end{equation}
where $[\cdot;\cdot]$ denotes sequence concatenation and $E(q)$ is omitted for captioning. The target embeddings $X \in \mathbb{R}^{L \times d}$ occupy $L$ output positions with embedding dimension $d$. They are corrupted to $Z_t$ according to Eq.~\eqref{eq:corruption}, where $t=0$ denotes the noise endpoint and $t=1$ the clean endpoint. The conditioning sequence remains uncorrupted.

We adapt a shared transformer with a denoising head and a token reconstruction head. In denoising mode, the model predicts clean embeddings from $Z_t$, conditioned on $C$ and a self-conditioning estimate $S$. In reconstruction mode, it predicts reference tokens from separately corrupted reference embeddings $\widetilde X$:
\begin{equation}
\widehat X_\theta = D_\theta(Z_t,t;C,S),\quad
\pi_\theta = R_\theta(\widetilde X,1;C).
\label{eq:visual_adaptation}
\end{equation}
Here, $S$ is a preliminary clean prediction computed without gradients, or zero when self-conditioning is omitted, and $\pi_\theta$ denotes the token distributions. Reconstruction uses a fixed decoder time input of $1$, irrespective of the corruption used to construct $\widetilde X$. This fixed decoder time input is omitted in subsequent notation. Bidirectional attention allows the shared transformer to jointly process the conditioning features and all output positions. At each update, we sample either the denoising or reconstruction mode, optimizing continuous denoising supervision or token-level cross-entropy, respectively.

We initialize the shared transformer and both output heads from pretrained ELF and train them jointly with $P_\phi$, keeping the image and text encoders frozen. To support subsequent image guidance, we apply image dropout by zeroing the projected visual features, yielding $C_{\varnothing}=[\mathbf{0};E(q)]$. This exposes the model to both image-present and image-zeroed conditioning. The corruption settings, scale-conditioning settings, and training-mode sampling probabilities are provided in the supplementary material.

\subsection{Stage II: Image Guidance Distillation}
\label{sec:guidance}

Classifier-free guidance combines image-conditioned and image-zeroed predictions to strengthen visual conditioning, but requires two denoiser evaluations per generation step. We distill this guidance into a single student prediction~\cite{meng2023distillation}. A frozen teacher $v_\psi$ and a trainable student $v_\theta$, including their respective visual projectors, are initialized from the adapted model (Figure~\ref{fig:architecture}(b), left). Here, $v_\psi$ and $v_\theta$ denote the velocity predictions obtained from clean-embedding estimates via Eq.~\eqref{eq:velocity}.

Let $C_\psi$ and $C_\theta$ denote the teacher and student conditioning sequences constructed as in Eq.~\eqref{eq:visual_condition}, using their respective projectors. The teacher additionally uses $C_\psi^{\varnothing}$, obtained by zeroing the projected image features while retaining the question conditioning. For a noisy text state $Z_t$ at time $t$, we compute
\begin{equation}
\begin{aligned}
v_{\mathrm{c}} &= v_\psi(Z_t,t;C_\psi,S),\\
v_{\mathrm{u}} &= v_\psi(Z_t,t;C_\psi^{\varnothing},S).
\end{aligned}
\label{eq:teacher_predictions}
\end{equation}
Here, $S$ is a preliminary clean prediction from the student, computed without gradients, or zero when self-conditioning is omitted. The student and both teacher evaluations share this self-conditioning input. The two teacher evaluations also use identical $Z_t$, $t$, and question conditioning, differing only in whether the explicit visual conditioning features are present or zeroed.

Given a guidance scale $w$, the teacher's guided velocity is
\begin{equation}
v_{\mathrm{g}}
=
v_{\mathrm{c}}+(w-1)(v_{\mathrm{c}}-v_{\mathrm{u}}).
\label{eq:guidance}
\end{equation}
For $w>1$, this amplifies the velocity difference induced by explicit image conditioning. The student predicts $v_\theta(Z_t,t;C_\theta,S,w)$ using image-present conditioning. We provide $w$ through the backbone's existing scale-conditioning interface, which adds an embedding of the scale to learned conditioning tokens. In this stage, the scale specifies the image guidance to be distilled. The student's denoising objective combines reference supervision with regression toward the teacher's guided velocity:
\begin{equation}
\begin{aligned}
\mathcal{L}_{\mathrm{student}}
&={}(1-\lambda)\mathcal{L}_{\mathrm{data}}\\
&+\lambda\,\mathbb{E}\!\left[
\left\|v_\theta(Z_t,t;C_\theta,S,w)
-\operatorname{sg}(v_{\mathrm{g}})\right\|_m^2
\right],
\end{aligned}
\label{eq:distillation}
\end{equation}
where $\lambda\in[0,1]$ controls the distillation weight, $\operatorname{sg}(\cdot)$ denotes stop-gradient, and $\|\cdot\|_m^2$ denotes masked mean squared error. The reference objective $\mathcal{L}_{\mathrm{data}}$ supervises the student velocity using the paired target embeddings, while the teacher term transfers the scale-dependent image guidance. Token reconstruction training is retained in the separate reconstruction mode. At inference, the student incorporates the distilled guidance through a single image-conditioned denoiser evaluation per step. Hereafter, $D_\theta$ denotes the distilled student's clean-embedding predictor.

\subsection{Stage III: Prediction-Adapted Decoding}
\label{sec:caption_reader}

Continuous generation ultimately requires mapping predicted embeddings to discrete tokens. However, the token reconstruction objective uses corrupted reference embeddings, whereas inference requires reading embeddings produced by the denoiser. Prediction errors can shift these inputs away from the representations encountered during reconstruction training, so accurate reconstruction of reference embeddings does not necessarily imply reliable readout of generated states. We therefore adapt the decoder directly to denoiser predictions, using their paired reference tokens as supervision.

To isolate readout adaptation from continuous generation, we initialize a separate decoder $R_\omega$ by copying the guided model's transformer and vocabulary head, while keeping the denoiser $D_\theta$ and visual projector frozen (Figure~\ref{fig:architecture}(b), right). The decoder retains multimodal conditioning $C$ and predicts all output tokens in parallel.

Given reference embeddings $X=E(y)$, we sample a timestep $t$ and obtain the noisy state $Z_t$ using Eq.~\eqref{eq:corruption}. The frozen denoiser first produces a preliminary clean estimate $\widehat X^{(0)}$, which is then used for self-conditioning in a second prediction:
\begin{equation}
\begin{aligned}
\widehat X^{(0)} &= D_\theta(Z_t,t;C,\mathbf{0},w),\\
\widehat X &= D_\theta(Z_t,t;C,\widehat X^{(0)},w).
\end{aligned}
\label{eq:decoder_prediction}
\end{equation}
Both evaluations use the same $Z_t$, $t$, conditioning $C$, and guidance scale $w$. The resulting $\widehat X$ provides the decoder input, while the reference tokens $y$ serve as cross-entropy targets. This exposes $R_\omega$ to denoiser predictions without requiring a complete generation trajectory.

We add reconstruction supervision on clean embeddings $X$ so that the decoder can also recover tokens directly from the text encoder representations. The decoder objective combines prediction-adapted decoding and clean-embedding reconstruction:
\begin{equation}
\begin{aligned}
\mathcal{L}_{\mathrm{dec}}
={}&(1-\lambda_{\mathrm{clean}})
\,\mathbb{E}\!\left[
\mathrm{CE}(R_\omega(\widehat X;C),y)
\right]\\
&+\lambda_{\mathrm{clean}}
\,\mathbb{E}\!\left[
\mathrm{CE}(R_\omega(X;C),y)
\right],
\end{aligned}
\label{eq:decoder_loss}
\end{equation}
where $\lambda_{\mathrm{clean}}\in[0,1]$ controls the contribution of clean-embedding reconstruction. Only the decoder parameters $\omega$ are updated, while $D_\theta$ and the visual projector remain frozen.

The resulting prediction-adapted decoder serves as the parallel token readout module at inference. For captioning, we also reuse this decoder during late denoising steps: its text predictions are re-encoded and fed back into continuous generation, as described in Section~\ref{sec:inference}. This feedback is applied with all model weights fixed and requires no additional decoder or denoiser training. The supplementary material provides the corruption ranges, guidance-scale settings, loss weights, and task-specific training details.

\subsection{Inference: Parallel Readout and Text Feedback}
\label{sec:inference}

Given multimodal conditioning $C$, generation starts from $Z_{t_0}=\sigma_{\mathrm{init}}\epsilon$, where $\epsilon\sim\mathcal{N}(0,I)$ and $\sigma_{\mathrm{init}}$ controls the inference initialization scale. The sampler jointly updates all $L$ output positions over $K$ denoising steps. At step $k$, the denoiser receives the current state concatenated with the previous clean estimate $S_k$, initialized as $S_0=\mathbf{0}$, and predicts $\widehat X_k=D_\theta(Z_{t_k},t_k;C,S_k,w)$. Image and question features remain fixed throughout generation.

We use the prediction-adapted decoder $R_\omega$ trained in Section~\ref{sec:caption_reader} for parallel token readout. For VQA, we use final-only readout: the decoder converts the final generated embeddings into text without modifying the preceding trajectory. For captioning, we additionally apply $R_\omega$ at selected late denoising steps and feed its text predictions back into the sampler through re-encoding. Let $\mathcal{T}$ denote parallel text readout from token logits. The clean estimate used by the sampler is
\begin{equation}
\bar X_k =
\begin{cases}
E\!\left(\mathcal{T}\!\left(R_\omega(\widehat X_k;C)\right)\right),
& k\in\mathcal{F},\\
\widehat X_k,
& \text{otherwise},
\end{cases}
\label{eq:inference_feedback}
\end{equation}
where $\mathcal{F}$ specifies the feedback steps and is empty for final-only readout. Before applying the normalized frozen text encoder $E$, the decoded text is retokenized and padded or truncated to $L$ positions, yielding an estimate in the denoiser's embedding space.

The sampler uses $\bar X_k$ to compute velocity via Eq.~\eqref{eq:velocity} and advance the noisy state, while setting $S_{k+1}=\bar X_k$ for subsequent self-conditioning. Thus, decoded text influences both the current update and later predictions without fixing a token prefix; all output positions remain editable.

After the final update, $R_\omega$ reads the resulting embeddings to produce the output text in parallel. All model weights remain fixed. This base inference configuration follows a single trajectory and uses the final readout without candidate selection. Training and inference noise scales, sampling settings, and implementation details are provided in the supplementary material.

%% file: sections/experiments_current.tex
\section{Experiments}
\label{sec:experiments}

We evaluate \method on visual question answering (VQA) and image captioning against autoregressive and masked diffusion baselines, exemplifying two existing paradigms for language modeling.

\subsection{Experimental Setup}

\noindent\textbf{Models and Baselines.} Our main experiments use \method with a language backbone initialized from pretrained ELF-B~\cite{hu2026elf}. We compare against autoregressive GPT-2~\cite{radford2019gpt2} and masked-diffusion MDLM~\cite{sahoo2024mdlm}, chosen for their comparable language-backbone sizes. All models use frozen SigLIP image features and the same downstream training data. To adapt each baseline, a learned visual projector maps the image features into its language embedding space and prepends them to the text input. We train the projector and fine-tune the language model using next-token prediction for AR and the masked-diffusion objective for MDLM. VQA baselines use greedy decoding, while captioning comparisons include both greedy and sampled outputs. Unless otherwise stated, \method uses 32 denoising steps, with text feedback during the final 10 steps. Hardware specifications and additional training and inference settings are provided in the supplementary material.

\vspace{2mm}
\noindent\textbf{Datasets and Evaluation.} For each baseline, we train separate models for VQA and image captioning. VQA models are trained on VQAv2~\cite{goyal2017vqa} and report soft-answer accuracy on a subset of its validation split. We further evaluate generalization to GQA~\cite{hudson2019gqa} and report exact-answer accuracy. All captioning comparisons and ablations use the COCO Karpathy split~\cite{lin2014coco}, reporting BLEU-4, CIDEr~\cite{vedantam2015cider}, and METEOR on the test set. The supplementary material details the dataset splits and image-overlap exclusions, as well as further captioning results on Flickr30k~\citep{young2014image}.

\subsection{Main Results}

We discuss our main results on VQA and image captioning. On VQA, Table~\ref{tab:vqa} demonstrates that \method outperforms both autoregressive and masked-diffusion baselines. On VQAv2, our method improves accuracy by 3.34 percentage points over AR and 8.47 points over MDLM. This advantage extends to GQA without further training: the same VQAv2-trained checkpoint exceeds AR by 2.83 points and MDLM by 4.50 points.

\begin{table}[htbp]
\centering
\caption{\textbf{\method outperforms baselines on VQAv2 and GQA.} \method uses 32 denoising steps with text feedback during the final 10 steps. Results are means over three inference seeds, with standard deviations in subscripts. MDLM and GPT-2 use greedy decoding and a single run.}
\label{tab:vqa}
\vspace{2mm}
\small
\setlength{\tabcolsep}{4pt}
\begin{tabular}{lccc}
\toprule
Model & Backbone (M) & VQAv2 (\%) & GQA (\%) \\
\midrule
MDLM     & 170 & 54.91 & 34.97 \\
GPT-2 AR & 124 & 60.05 & 36.64 \\
\method  & 105 & $\textbf{63.39}_{\pm0.03}$ & $\textbf{39.47}_{\pm0.40}$ \\
\bottomrule
\end{tabular}
\end{table}

\noindent Table~\ref{tab:caption} compares \method with autoregressive and masked-diffusion baselines on image captioning. Because an image can admit multiple valid descriptions, we evaluate both greedy decoding and stochastic sampling. \method achieves 111.16 CIDEr and outperforms sampled AR and both MDLM decoding settings across all three metrics; it is furthermore competitive with greedy AR's 114.66 CIDEr, with similar BLEU-4 and METEOR scores.

\begin{table}[htbp]
\centering
\caption{\textbf{\method outperforms sampled AR on COCO Karpathy test.} Sampled baselines use temperature 0.7 and top-$p$ 0.95. Both \method and MDLM use 32 denoising steps; \method uses text feedback in the last ten steps. Subscripts indicate standard deviations over three inference seeds; greedy results use a single run. All metrics are reported on a $\times 100$ scale.}
\label{tab:caption}
\vspace{2mm}
\small
\setlength{\tabcolsep}{3pt}
\renewcommand{\arraystretch}{1.15}
\begin{tabular}{@{}lccc@{}}
\toprule
Model & BLEU-4 $\uparrow$ & CIDEr $\uparrow$ & METEOR $\uparrow$ \\
\midrule
\method & $33.00_{\pm0.45}$ & $111.16_{\pm0.69}$ & $27.34_{\pm0.08}$ \\
\quad + Retrieval & $\mathbf{33.40}_{\pm0.29}$ & $\mathbf{114.69}_{\pm0.81}$ & $\mathbf{27.47}_{\pm0.08}$ \\
\midrule
\multicolumn{4}{@{}l}{\textit{Greedy decoding}} \\[2pt]
MDLM & 27.60 & 94.91 & 23.82 \\
GPT-2 AR & 33.29 & 114.66 & 27.36 \\
\midrule
\multicolumn{4}{@{}l}{\textit{Sampling: temperature 0.7, top-$p$ 0.95}} \\[2pt]
MDLM & $27.15_{\pm0.14}$ & $93.44_{\pm0.11}$ & $23.80_{\pm0.03}$ \\
GPT-2 AR & $25.04_{\pm0.41}$ & $93.35_{\pm0.39}$ & $25.08_{\pm0.12}$ \\
\bottomrule
\end{tabular}
\end{table}

\paragraph{Retrieval-based draft selection.} 
Unlike left-to-right decoding, whose intermediate states are incomplete prefixes, every \method readout is a complete caption, so a trajectory provides alternative drafts without extra samples or beams. We augment intermediate readouts with single-word variants and rank the candidates by similarity-weighted CIDEr-D against captions of the 20 nearest training images, retrieved using cosine similarity of SigLIP image embeddings. We replace the final readout only when the score improves by at least 2, with settings selected on validation; details are provided in the supplementary material.

\paragraph{Effect of Image Guidance.} Our distillation stage uses a frozen teacher to provide image-guided denoising targets, which the generator learns to match using a single image-conditioned prediction at inference. To test the benefit of these targets, we compare caption training with and without the distillation term in Eq.~\eqref{eq:distillation}. Both conditions start from the same checkpoint after initial image-conditioned adaptation and use identical data, update budgets, and learning-rate schedules. Before token decoder adaptation, guided training raises CIDEr from 92.05 to 96.97 (Table~\ref{tab:matched_guidance}). The 4.93-point gain under matched training budgets supports distilling image guidance into the generator rather than relying on additional training alone.

\begin{table}[htbp]
\centering
\caption{\textbf{Guided targets improve captioning under matched training budgets.} Only regression to the teacher's guided prediction changes. Both models use 32 denoising steps without token decoder adaptation or text feedback on Karpathy test. Scores are means over three seeds, with standard deviations in subscripts.}
\label{tab:matched_guidance}
\vspace{2mm}
\small
\setlength{\tabcolsep}{4pt}
\begin{tabular}{lccc}
\toprule
Continuation & BLEU-4 $\uparrow$ & CIDEr $\uparrow$ & METEOR $\uparrow$ \\
\midrule
Without guided target & $26.80_{\pm0.14}$ & $92.05_{\pm0.37}$ & $25.64_{\pm0.07}$ \\
With guided target & $27.28_{\pm0.11}$ & $96.97_{\pm0.22}$ & $27.09_{\pm0.09}$ \\
\bottomrule
\end{tabular}
\end{table}

\paragraph{Token Decoder Adaptation.} We compare two ways of constructing decoder-training inputs from training captions: denoiser predictions and noisy reference embeddings. Both conditions use the same reference-text supervision, decoder architecture, and update budget, with the guided denoiser fixed. We also evaluate the decoder without adaptation (Table~\ref{tab:decoder_controls}).

Training on denoiser predictions yields 63.39\% VQAv2 accuracy, compared with 62.48\% for noisy-reference training and 62.52\% without adaptation. The 0.87-point gain over the matched noisy-reference control supports training the decoder to recover text from denoiser predictions rather than from artificially corrupted reference embeddings.

\begin{table}[htbp]
\centering
\caption{\textbf{Training the token decoder on denoiser predictions improves VQA.} The denoiser is fixed, and the trained decoders share the same architecture and update budget. This component comparison uses eight denoising steps without text feedback. Scores are means over three inference seeds, with standard deviations in subscripts.}
\label{tab:decoder_controls}
\vspace{2mm}
\small
\setlength{\tabcolsep}{4pt}
\begin{tabular}{lc}
\toprule
Decoder adaptation & VQAv2 (\%) $\uparrow$ \\
\midrule
None & $62.52_{\pm0.03}$ \\
Noisy references & $62.48_{\pm0.03}$ \\
Denoiser predictions & $63.39_{\pm0.03}$ \\
\bottomrule
\end{tabular}
\end{table}

The same comparison on captioning also favors training on denoiser predictions (Table~\ref{tab:caption_decoder_controls}). With the generator fixed, prediction-adapted decoding reaches 107.17 CIDEr, compared with 104.79 for noisy-reference training and 104.90 without adaptation. The 2.38-point gain over the matched noisy-reference control shows that the benefit extends beyond short answers to caption generation.

\begin{table}[htbp]
\centering
\caption{\textbf{Prediction-adapted decoding improves captioning.} The generator is fixed; trained decoders share the same architecture and update budget. All rows use 32 denoising steps without text feedback on COCO Karpathy test. Subscripts give standard deviations over three inference seeds.}
\label{tab:caption_decoder_controls}
\vspace{2mm}
\small
\setlength{\tabcolsep}{4pt}
\begin{tabular}{@{}lccc@{}}
\toprule
Decoder adaptation & BLEU-4 & CIDEr & METEOR \\
\midrule
None & $30.57_{\pm0.24}$ & $104.90_{\pm0.46}$ & $27.22_{\pm0.08}$ \\
Noisy references & $30.57_{\pm0.26}$ & $104.79_{\pm0.49}$ & $27.19_{\pm0.07}$ \\
Denoiser predictions & $31.09_{\pm0.14}$ & $107.17_{\pm0.25}$ & $27.37_{\pm0.05}$ \\
\bottomrule
\end{tabular}
\end{table}

\paragraph{Effect of Text Feedback.} Text feedback re-encodes the adapted token decoder's intermediate predictions to refine the late denoising updates. With model weights and 32 denoising steps fixed, the feedback procedure raises CIDEr from 107.17 to 111.16 and BLEU-4 from 31.09 to 33.00, while METEOR remains similar (Table~\ref{tab:text_feedback}). This 3.99-point CIDEr gain requires no additional training. The comparison measures the complete feedback procedure, including reduced late-step re-noising; it does not attribute the entire gain to text re-encoding alone.

\begin{table}[htbp]
\centering
\caption{\textbf{Text feedback improves captioning without further training.} Both rows use the same model, 32 steps, initial noise and final phrase cleanup on Karpathy test. The second row adds text feedback and reduced late-step re-noising. Subscripts give standard deviations over three inference seeds.}
\label{tab:text_feedback}
\vspace{2mm}
\small
\setlength{\tabcolsep}{4pt}
\begin{tabular}{@{}lccc@{}}
\toprule
Model & BLEU-4 $\uparrow$ & CIDEr $\uparrow$ & METEOR $\uparrow$ \\
\midrule
Without feedback & $31.09_{\pm0.14}$ & $107.17_{\pm0.25}$ & $27.37_{\pm0.05}$ \\
With feedback & $33.00_{\pm0.45}$ & $111.16_{\pm0.69}$ & $27.34_{\pm0.08}$ \\
\bottomrule
\end{tabular}
\end{table}

As Table~\ref{tab:answer_feedback} shows, text feedback provides a further VQAv2 gain after token decoder adaptation. With the denoiser, token decoder, and noise settings fixed, accuracy increases by 0.42 percentage points, with a paired 95\% confidence interval of $[0.37,0.46]$. GQA accuracy remains nearly unchanged. The feedback benefit is therefore task-dependent: it improves VQAv2 answering but does not provide a measurable transfer gain on GQA.

\begin{table}[htbp]
\centering
\caption{\textbf{Effect of text feedback on VQA.} The denoiser, token decoder, 32 steps, noise settings and removal of adjacent repetitions are fixed. Only text feedback changes. Scores are means over three inference seeds, with standard deviations in subscripts.}
\label{tab:answer_feedback}
\vspace{2mm}
\small
\setlength{\tabcolsep}{4pt}
\begin{tabular}{@{}lcc@{}}
\toprule
Inference & VQAv2 (\%) $\uparrow$ & GQA (\%) $\uparrow$ \\
\midrule
Without text feedback & $62.97_{\pm0.01}$ & $39.46_{\pm0.21}$ \\
With text feedback & $63.39_{\pm0.03}$ & $39.47_{\pm0.40}$ \\
\bottomrule
\end{tabular}
\end{table}

\subsection{Number of Denoising Steps}

We vary the number of denoising steps with text feedback disabled, keeping the denoiser and adapted token decoder fixed. As Table~\ref{tab:step_count} shows on image captioning, CIDEr is 107.17 at the default 32-step budget, compared with 104.50 at 16 steps and 106.02 at 64 steps. METEOR increases from 26.80 at 16 steps to 27.44 at 64 steps, so additional denoising does not uniformly improve all metrics.

\begin{table}[htbp]
\centering
\caption{\textbf{Effect of denoising steps on captioning without text feedback.} The denoiser and token decoder are fixed. Evaluation uses COCO Karpathy test, with means and standard deviations over three inference seeds.}
\label{tab:step_count}
\vspace{2mm}
\small
\setlength{\tabcolsep}{4pt}
\renewcommand{\arraystretch}{1.1}
\begin{tabular}{@{}rrr@{}}
\toprule
Steps & CIDEr & METEOR \\
\midrule
16 & $104.50_{\pm0.26}$ & $26.80_{\pm0.09}$ \\
32 & $107.17_{\pm0.25}$ & $27.37_{\pm0.05}$ \\
64 & $106.02_{\pm0.19}$ & $27.44_{\pm0.03}$ \\
\bottomrule
\end{tabular}
\end{table}

\subsection{Analysis and Discussion}
\label{sec:experimental_analysis}

Tables~\ref{tab:matched_guidance} and~\ref{tab:decoder_controls} isolate the contributions of image guidance and token decoder adaptation. Guided targets improve captioning under matched training budgets, while training the decoder on denoiser predictions improves VQA with the denoiser fixed. Both comparisons retain reference supervision and the same update budget, so additional training alone cannot explain the gains. The decoder comparison further demonstrates that adapting the mapping from continuous predictions to text can improve output quality without further denoiser training.

Text feedback extends the adapted decoder's role beyond final token recovery by using its intermediate readouts to refine subsequent denoising updates (Tables~\ref{tab:text_feedback} and~\ref{tab:answer_feedback}). For captioning, the feedback procedure improves output quality with model weights and the 32-step denoising budget held fixed. This procedure combines re-encoding of decoder predictions with reduced late-step re-noising. The answering comparison keeps the noise settings unchanged and therefore isolates the contribution of text feedback. These results show how the independently adapted decoder can improve both the final readout and the trajectory that produces it.

\subsection{Qualitative Results}

\begin{figure}[H]
\centering
\includegraphics[width=0.72\linewidth]{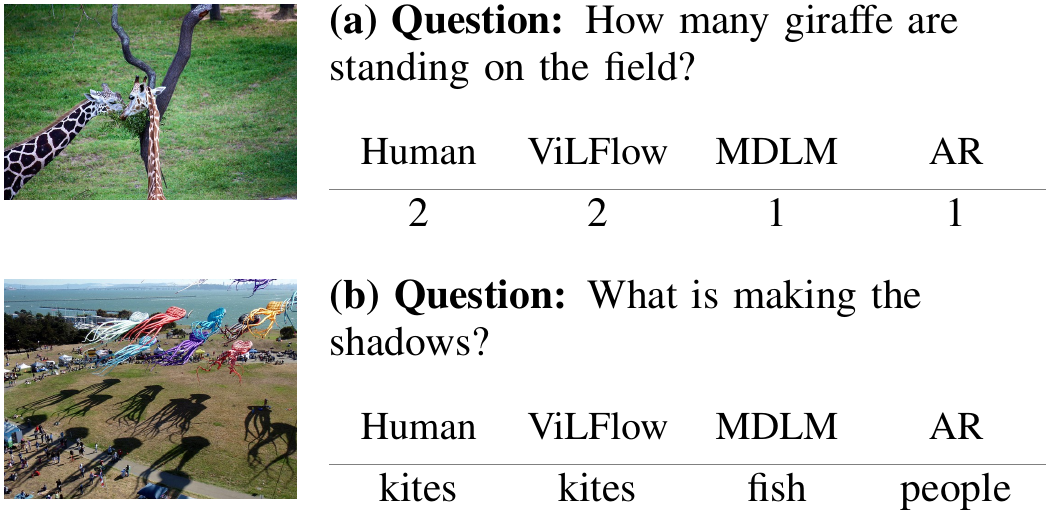}
\caption{\textbf{VQA examples.} Human denotes the most frequent annotated answer. The examples use the ablation models without text feedback; AR and MDLM decode greedily.}
\label{fig:qualitative_vqa}
\end{figure}

Figure~\ref{fig:qualitative_vqa} illustrates counting and scene interpretation. \method correctly counts the two overlapping giraffes in (a) and identifies the kites casting the shadows in (b), while both baselines give incorrect answers.

\begin{figure}[H]
\centering
\includegraphics[width=0.72\linewidth]{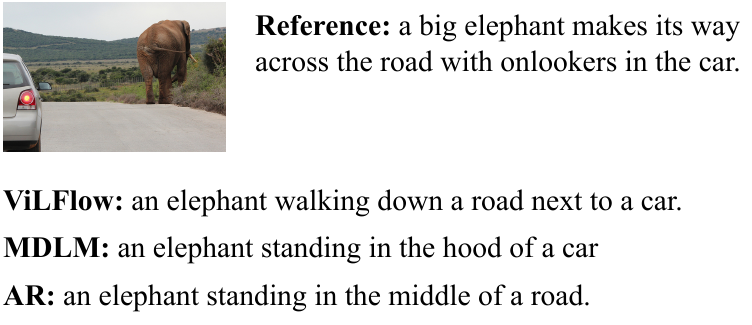}
\caption{\textbf{COCO captioning example.} \method describes the elephant's position relative to the car. Outputs use the ablation models without text feedback; AR and MDLM decode greedily.}
\label{fig:qualitative_caption}
\end{figure}

Figure~\ref{fig:qualitative_caption} compares descriptions of the same scene. \method describes the elephant next to the car. AR omits the car, whereas MDLM places the elephant in its hood. This comparison illustrates a difference in the described spatial relation, not just the objects mentioned.

\FloatBarrier

%% file: sections/conclusion.tex
\section{Conclusion}
\label{sec:conclusion}

We present \method, a continuous vision–language model for question answering and captioning. Our results show that visual adaptation benefits from both learning image guidance and training the token decoder to interpret denoiser predictions. Decoder adaptation improves both tasks without further denoiser training, while text feedback further improves captioning. These findings establish token decoding as more than a final conversion into words: it is an independently trainable component that can also improve continuous generation. Our evaluation focuses on short answers and captions; extending the approach to longer, open-ended visual responses remains future work.